\documentclass[11pt,a4paper]{article}
\usepackage[hyperref]{acl2021}
\usepackage{times}
\usepackage{latexsym}

\usepackage{microtype}
\usepackage{graphicx}
\aclfinalcopy 

\title{Prompt-tuning in ASR systems for efficient domain-adaptation}

\author{Saket Dingliwal\thanks{*equal contribution} \,\,\,  Ashish Shenoy\footnotemark[1] \,\,\, Sravan Bodapati \\ \textbf{Ankur Gandhe \,\,\, 
Ravi Teja Gadde \,\,\, Katrin Kirchhoff} \\
\{skdin, ashenoy, sravanb, aggandhe, gadderav, katrinki \}@amazon.com}

\date{}
\usepackage{lipsum}

\newcommand\blfootnote[1]{%
  \begingroup
  \renewcommand\thefootnote{}\footnote{#1}%
  \addtocounter{footnote}{-1}%
  \endgroup
}

\begin{document}
\maketitle

\blfootnote{Appearing at WeCNLP (West Coast NLP) Summit 2021}

\section{Introduction}
Automatic Speech Recognition (ASR) systems form a key component of various products across industry. Many of these ASR systems rely on a complex Acoustic Model (AM) whose output is rescored by a domain-specific Language Model (LM).  
As we use ASR systems in new domains, the memory, maintenance and data-collection costs for these domain-specific LMs increase.
Particularly, with advent  of parameter-heavy Transformer based LMs \citep{devlin-etal-2019-bert}, maintaining multiple domain-specific LMs is practically infeasible. While on the other hand, using a generic LM for all domains falls short in performance when compared to multiple domain-specific LMs. Therefore, a need for a middle ground between performance and costs is evident. 

To overcome this problem, we bring forward a methodology based on recently proposed Prompt Tuning. \citet{lester2021power} introduced this idea of learning the token embeddings of the prompt used to prime a LM to a particular task. Prompts are special tokens describing a task which when appended to the input data sample, helps the model understand and use this problem description to better solve the task. For example, to solve the machine translation task, instead of fine-tuning the Transformer model with corresponding dataset, one can achieve comparable performance by just showing text describing machine translation to the  powerful Transformer-based LM. In prompt tuning, instead of providing this prompt manually to the model, one learn it from the labelled examples from the task.

To the best of our knowledge, we are the first one to apply prompt-tuning for domain-adaptation of ASR systems. We find that one can learn the prompt-embeddings for not only different tasks but also different domains of the same task. This means there exists a domain description, which when passed to model as prefix, helps it better score sentences from that domain. We hypothesize that this is due to the fact that self-attention in Transformer model will create a positive interaction between domain embeddings (description) and the words from sentence, resulting in better prediction of the next token. This has an important usecase in multi-domain ASR systems, where learning a small number of parameters corresponding to domain prompt embedding can save costs associated with maintaining separate copies of LMs for each domain. Particularly, with Transformer models with millions of parameters, fine-tuning can be very slow and requires a large amount of domain data to prevent overfitting. However, with prompt-tuning one can achieve similar performance with <0.1\% of the parameters required for fine-tuning. We provide experimental evidence by comparing the performance of prompt-tuned and fine-tuned versions of LMs to score sentences from different domains of a dialog dataset. We also provide ablations on different sizes of prompt.

\section{Experiments}

\begin{figure*}
    \centering
    \includegraphics[width=0.8\textwidth]{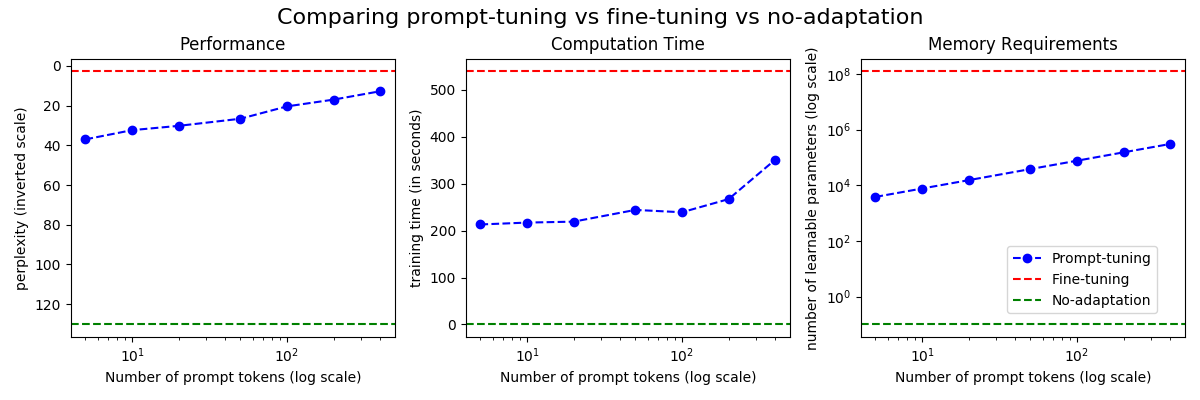}
    \caption{Comparing the performance, computation and memory costs of different methods of domain adaptation}
    \label{fig:comp}
\end{figure*}

We design first set of experiments as a proof of concept to show that prompt-tuning works for domain adaptation of LM. For this, we used textual data from different domains of the publically available MultiDoGo dataset \citep{peskov-etal-2019-multi}. We split the data from each domain into train, dev and test. Then for each domain individually, we fine-tune (and prompt-tune) a pretrained GPT-2 \citep{radford2019language} using the train set, tune the hyper-parameters using dev set and compute the perplexity numbers on the test set. We also evaluate the test set for each domain using the same out-of-the-box pretrained GPT-2. We then compare the no-adaptation, fine-tuned and prompt-tuned versions for each domain for different performance, memory and computation time metrics. Rather than fixing the size of the prompt, we repeat the above experiment with different sizes to show how we can control the trade-offs between performance and costs using this hyper-parameter. Although, we summarize the results for one such domain (fastfood) in Fig. \ref{fig:comp}, the results are consistent across different domains. All the experiments were conducted using a machine with 16 Tesla K80 GPUs and the reported times are wall clock times in seconds. As we can see in Fig. \ref{fig:comp}, with as little as 5 prompt-tokens, the perplexity improves from 130 in out-of-the-box model (green line) to 37. With 400 prompt-tokens, we achieve a very similar perplexity to fully fine-tuned model (red line) but with number of learnable parameters reducing by large orders of magnitude. As one might expect, with increase in size of prompts, the performance improves but so do the memory and computation costs. 

The above experiment proves us the effectiveness of prompt-tuning in domain adaptation. With comparable perplexity as fine-tuning, it is natural to expect that prompt-tuned model can perform at par with fully tuned model in domain specific rescoring of n-best hypothesis from an AM model in any ASR system. We verify this empirically by using the above models as first pass rescorers on an internal fastfood audio dataset for a hybrid AM model and measuring the Content Word Error Rates (CWER). While the fine-tuned version brings a relative improvement of 7.3\% over the no-adaptation baseline, training just 10 prompt tokens could already achieve 5.8\% relative improvement. Also, with 400 prompt tokens, we could almost exactly match the CWER of fine-tuned model. Since, ASR systems are used in tens of different domains, one can achieve a significant reduction in LMs that one need to maintain, by keeping a single copy of the base version of the model and saving the domain-specific prompt token embeddings. 

\section{Future Work}
While the initial experiments gives us confidence about the potential use of prompt-tuning in LMs in ASR systems, we plan to explore a lot more as part of this ongoing work. We will be running multiple experiments to showcase the benefits in WER further in multiple domains, datasets, etc. While in our experiments, we used the complete training data to fine-tune (and prompt-tune) the models, we believe in limited domain data setting, prompt-tuned models might even outperform the corresponding fine-tuned versions as the number of learnable parameters are less. This will be especially critical for new domains in ASR where getting high-quality data is difficult. We initialized our prompt token embedding layer randomly but as suggested in \citep{lester2021power}, we plan to run an ablation on different choices of initializations. One such promising initialization can be embedding for the most frequent words in that domain. In addition to all these experiments, we also plan to compare our version of domain-adaptation with related works of AdapterHub \citep{pfeiffer-etal-2020-adapterhub}, meta-learning \citep{bansal-etal-2020-learning}, using semantic embeddings \citep{shenoy21_interspeech, shenoy2021asr} and freezing all but last layer of Transformer model. The analysis provided in this paper will act as a proof of concept on computation and memory efficiency of prompt-tuning guiding  our future directions.

\bibliographystyle{acl_natbib}
\bibliography{anthology,acl2021}


\end{document}